\documentclass{article}

\PassOptionsToPackage{numbers, compress}{natbib}
    \usepackage[preprint]{neurips_2024}

\usepackage{multirow}
\usepackage[utf8]{inputenc} % allow utf-8 input
\usepackage[T1]{fontenc}    % use 8-bit T1 fonts
\usepackage{hyperref}       % hyperlinks
\hypersetup{
    colorlinks=true,
    linkcolor=black,    % 내부 링크 색상
    citecolor=black,    % 인용문 색상
    urlcolor=black      % URL 색상
}
\usepackage[numbers]{natbib}
\usepackage{url}            % simple URL typesetting
\usepackage{booktabs}       % professional-quality tables
\usepackage{amsfonts}       % blackboard math symbols
\usepackage{nicefrac}       % compact symbols for 1/2, etc.
\usepackage{microtype}      % microtypography
\usepackage{xcolor}         % colors
\usepackage{fancyhdr}       % header
\usepackage{graphicx}       % graphics
\usepackage[inkscapelatex=false]{svg}
\usepackage{amsmath}
\usepackage{wrapfig}
\usepackage{subcaption}  % For subfigures
\usepackage{float}

\usepackage{bbm}
\usepackage[normalem]{ulem}

\title{Amnesia as a Catalyst for Enhancing Black Box Pixel Attacks in Image Classification and Object Detection}

\author{%
  Dongsu Song \\
  Department of Artificial Intelligence\\
  Korea Aerospace University\\
  \texttt{raister2873@gmail.com}\\
  \And
  Daehwa Ko \\
  Department of Software\\
  Korea Aerospace University\\
  \texttt{daehwa001210@gmail.com}
  \And
  Jay Hoon Jung\thanks{Corresponding author.} \\
  Department of Artificial Intelligence\\
  Korea Aerospace University\\
  \texttt{jhjung@kau.ac.kr}\\
 }

\begin{document}

\maketitle

\begin{abstract}
    It is well known that query-based attacks tend to have relatively higher success rates in adversarial black-box attacks.
    While research on black-box attacks is actively being conducted, relatively few studies have focused on pixel attacks that target only a limited number of pixels.
    In image classification, query-based pixel attacks often rely on patches, which heavily depend on randomness and neglect the fact that scattered pixels are more suitable for adversarial attacks.
    Moreover, to the best of our knowledge, query-based pixel attacks have not been explored in the field of object detection.
    To address these issues, we propose a novel pixel-based black-box attack called \textbf{R}emember and \textbf{F}orget \textbf{P}ixel \textbf{A}ttack using \textbf{R}einforcement Learning(RFPAR), consisting of two main components: the Remember and Forget processes.
    RFPAR mitigates randomness and avoids patch dependency by leveraging rewards generated through a one-step RL algorithm to perturb pixels.
    RFPAR effectively creates perturbed images that minimize the confidence scores while adhering to limited pixel constraints.
    Furthermore, we advance our proposed attack beyond image classification to object detection, where RFPAR reduces the confidence scores of detected objects to avoid detection. 
    Experiments on the ImageNet-1K dataset for classification show that RFPAR outperformed state-of-the-art query-based pixel attacks.
    For object detection, using the MS-COCO dataset with YOLOv8 and DDQ, RFPAR demonstrates comparable mAP reduction to state-of-the-art query-based attack while requiring fewer query.
    Further experiments on the Argoverse dataset using YOLOv8 confirm that RFPAR effectively removed objects on a larger scale dataset.
    Our code is available at \href{https://github.com/KAU-QuantumAILab/RFPAR}{https://github.com/KAU-QuantumAILab/RFPAR.}

\end{abstract}

\section{Introduction}
\label{intro}
Deep learning models are susceptible to adversarial attacks, which involve subtle modifications of input data that are imperceptible to humans but lead to incorrect predictions by the model\cite{szegedy2014intriguing}.
As deep learning technologies become commercialized in the real world, the issue of adversarial attacks has garnered increasing attention.

Adversarial attacks can be broadly categorized into white-box attacks and black-box attacks\cite{bhambri2020survey}. In white-box attacks\cite{fgsm, ffgsm, pgd}, attackers devise attack strategies based on internal information about deep learning models, such as training data, gradients of the outputs with respect to the weights, and other details about the learning process for given samples.
Conversely, in black-box attacks\cite{black1,black2}, attackers can access only limited information such as the probability of the correct prediction for a given sample.
Given that real-world attackers typically only possess limited information about the model, black-box attacks are more realistic than white-box attacks.
In other words, research on black-box attacks and their defenses is crucial in order to develop robust and secure machine learning systems. 

\begin{figure}
    \centering
    % \includesvg[width=1\linewidth]{figure/fig1_white.svg}
    \includegraphics[width=1\textwidth]{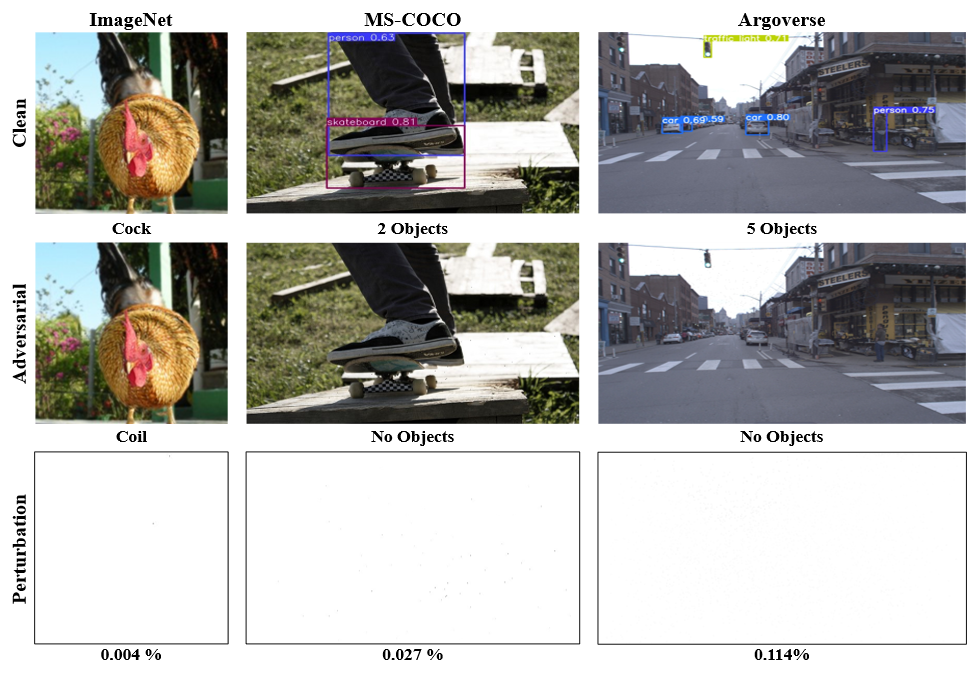} 
    \caption{\textbf{Adversarial examples generated by RFPAR.} The first column represents images from ImageNet (image classification), the second column from MS-COCO (object detection), and the third column from Argoverse (object detection). Each row represents a different condition: the first row shows clean images, the second row shows adversarially perturbed images, and the third row shows the perturbation levels with the ratio of attacked pixels to total pixels. Labels in the images indicate detected objects or classifications, such as "Cock" in ImageNet, "2 Objects" in MS-COCO, and "5 Objects" in Argoverse. In the adversarial row, labels are altered due to perturbations, resulting in misclassifications or undetected objects, such as "Coil" instead of "Cock" in ImageNet and no objects detected in MS-COCO and Argoverse. The perturbation row indicates the percentage of pixels attacked in the image. The percentages were 0.004\% for ImageNet, 0.027\% for MS-COCO, and 0.114\% for Argoverse.}
    \label{fig:fig1}
\end{figure}
Black-box attacks are also categorized into query-based methods\cite{Onepixel,Scratch,RLAB,Pixel} and transfer-based methods\cite{trasnfer}. 
Query-based attacks are generating adversarial examples by repeatedly querying the victim model with modified images\cite{query_expla}.
Transfer-based attacks involve generating adversarial examples for a surrogate model that successfully deceive another model\cite{transfer_explain}.
Transfer-based attacks are highly efficient since they do not require knowledge of the victim model. However, the discrepancies in model architecture, training data, and training methodologies between the surrogate and victim models often result in a lower success rate for these attacks compared to query-based attacks\cite{query_expla}. Conversely, although query-based attacks achieve higher success rates, they require a significant number of queries to the victim model. Therefore, reducing the number of queries in query-based attacks is a critical issue. 

The pioneering pixel attack method, OnePixel\cite{Onepixel}, employed Differential Evolution (DE) to generate adversarial images. An advanced approach, ScratchThat\cite{Scratch}, used DE to create curves and applied a parametric model to perturbations, reducing parameters and improving performance. A more recent
study, PIXLE\cite{Pixel}, enhanced query efficiency and attack success rate by using a simple algorithm instead of DE. Briefly, PIXLE generates adversarial images by selecting arbitrary patches in a clean image and applying the brightness of these pixels to others.
Although this method improved performance, it ignored the fact that pixels are independent of each other due to its reliance on patches and exhibited inefficiencies stemming from randomness in brightness mapping.
The previous study, PatchAttack\cite{Patchattack}, utilized RL model to embed textures in specific regions of the clean image, discovering vulnerable patches and reducing randomness, which significantly decreased the number of queries and improved attack success rates. Unfortunately, this method still depended on patches, requiring at least 3\% of the image area to be attacked. 

\begin{figure}
    \centering
    % \includesvg[width=1\linewidth]{figure/fig2_new.svg}
    \includegraphics[width=1\textwidth]{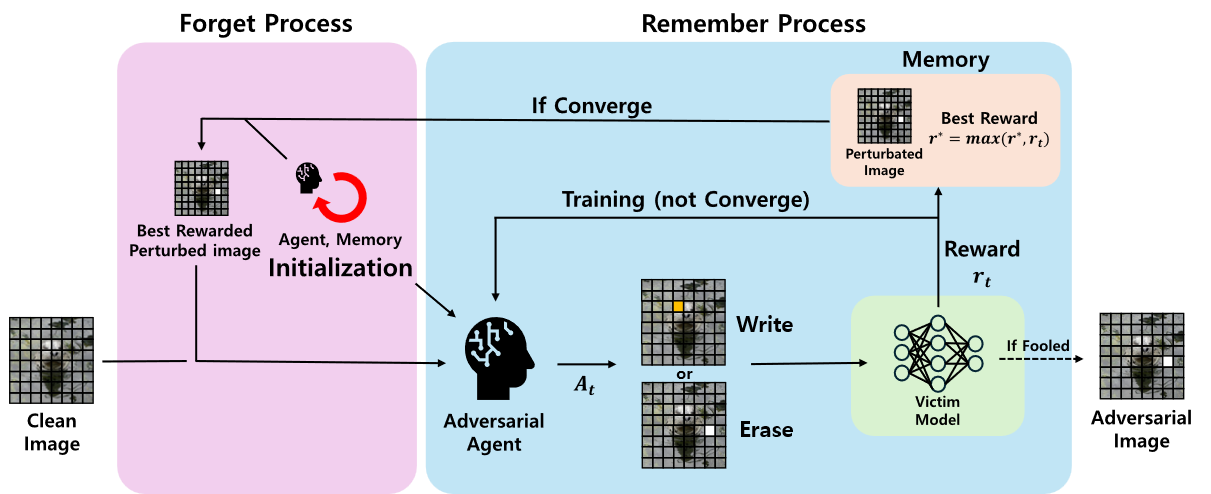} 
    \caption{The model architecture of RFPAR: the Remember and Forget process. During the Remember process, the RL model generates perturbed images and corresponding rewards. Memory compares these with previously stored values and retains only the highest reward and its associated image. 
    Once the rewards converge to a certain value, the Forget process starts and resets the RL agent and memory, then reintroduces the perturbed images that gained the highest reward to the Remember process.
    The process continues until an adversarial image is generated or a predefined number of cycles is reached, at which point it terminates.}
    \label{fig:fig2}
\end{figure}

Query-based attacks in object detection are more challenging than those in image classification. The first query-based attack in object detection, PRFA\cite{PRFA}, generated adversarial images using a parallel rectangle flipping strategy. Recent research, GARSDC\cite{GARSDC}, employed a genetic algorithm to create adversarial images, improving optimization efficiency by using adversarial examples generated from transfer-based attacks as the initial population. Query-based attacks on black-box models are inherently challenging, and targeting only a few pixels is even more difficult to study. To the best of our knowledge, pixel attacks have been limited to white-box or transfer attack methods\cite{od_pixel1,od_pixel2,od_pixel3,od_pixel4}. In this study, we extend our proposed attack from image classification to object detection, introducing the first query-based pixel attack. Experiments show that our method achieves a comparable mAP reduction on YOLO\cite{YOLOv8} to state-of-the-art methods while significantly reducing the number of queries, demonstrating its effectiveness in object detection.

In this study, we introduce a novel method called the Remember and Forget Pixel Attack using Reinforcement Learning (RFPAR). Briefly, in the Remember process, the clean image is initially taken as input by the RL agent, and the loss function is optimized.
During this optimization, the highest reward and its corresponding perturbed image are stored in memory.
If the highest rewards do not change for a while, we define this as the convergence of rewards.
Once the rewards converge, the Forget process is initiated, resetting the RL agent and memory to forget previous information. 
After resetting, the stored image is fed as input to the RL agent, and the Remember process begins again.
Extensive experiments demonstrate that our attack is effective for image classification and successfully extends to object detection.

In summary, our main contributions are:

\begin{itemize}
\item
We propose a novel query-based black-box pixel attack consisting of the Remember and Forget processes.
Our approach outperformed state-of-the-art attacks on the ImageNet-1K classification task, achieving an average attack success rate improvement of 12.1\%, while reducing the number of queries by 26.0\% and the modified $L_0$ norm by 41.1\%.

\item We advance query-based pixel attacks from image classification to object detection, introducing the query-based pixel attack.
Our experiments demonstrate that our proposed method effectively compromises object detection systems.
It achieves an average mean Average Precision (mAP) reduction of 0.29 in YOLO, comparable to state-of-the-art query-based attacks, while reducing the number of queries by 52.8\%. 
To the best of our knowledge, the proposed method is the first black-box query-based pixel attack for object detection.

\item To evaluate performance on a larger scale dataset, we conducted experiments using YOLOv8 as the victim model on the Argoverse-1.1 validation dataset.
We also achieved a high removal rate of detected objects above 0.9 in Argoverse, similar to our results in MS-COCO.
The results demonstrate that our proposed method effectively reduces the number of detected objects in images with a resolution of 1920$\times$1200.
Additionally, examining the adversarial perturbation results of RFPAR on Argoverse and MS-COCO, we observe that it successfully reduces the number of detected objects while attacking only very small areas of 0.1\% and 0.02\%, respectively.
\end{itemize}

\section{Remember and Forget Pixel Attack Using Reinforcement Learning}
\label{RFPAR}
In this section, we introduce our proposed method. In Section \ref{formalization}, we define the problem mathematically. Section \ref{Remember Process} details the Remember process, which is the internal iterative structure of our algorithm, including agent, environment, and memory. Finally, we explain the Forget process, which serves as the external iterative structure in Section \ref{forget process}. 

\subsection{The Problem Formalization}
\label{formalization}
We consider an image classifier as \(f: \mathbb{R}^{C\times H\times W} \rightarrow \mathbb{R}^{c}  \), where  $C$, $H$, and $W$ represent the channel, height, and width of a given sample $x$, respectively, and \( c \) denotes the number of classes. 
The classifier $f$ computes probabilities for each class for a sample $x$.
Furthermore, $f_l (x)$ is defined as the probability of a sample $x$ being predicted as the $l$-th class.
The prediction of the classifier can be expressed as ${\arg\max}_l f_l(x)$.
For an image classification attack, the objective is to minimize a pixel perturbation $\delta$ such that the classifier's prediction for an input \( x \) diverges from its true label \( y \).
This problem can be formalized as:

\begin{equation}
\begin{aligned}
\underset{\delta}{\min} \quad  & \underset{l}{\arg\max}   \, f_l(x+\delta) \neq y \\
\text{s.t.} \quad & \|\overline{x} - x\|_0 = \|\delta\|_0 \leq \epsilon, \quad \epsilon \in \mathbb{Z}^+.
\end{aligned}
\end{equation}

Here, $\epsilon \in \{1,2,\cdots\}$ and $\left\| \cdot \right\|_0$ denotes the attack level and the $L_0$ norm. $\overline{x}$ is the perturbed image, defined as $x+\delta$. Notably, the attack levels are positive integers, which implies that the perturbations are at the pixel level.

Similarly, the object detector is defined by the function \( f: \mathbb{R}^{C \times H \times W} \rightarrow \mathbb{R}^{B \times 6} \), where \( B \) represents the maximum number of objects that can be detected by the model.
The information about each bounding box location, object's class, and confidence score (indexed by $l_0$) is encapsulated in 6-dimensional vectors.
To prevent the attack from targeting objects that are incorrectly predicted, we establish a confidence threshold of 0.5.
If $n$ objects surpass this threshold from among $B$ candidates, then only these $n$ objects are classified as detected. 
Furthermore, \( f_{l_0}^o(x) \) indicates the confidence score that the model identifies the \( o \)-th object as belonging to the predicted class from a sample \( x \).
In this context, the number $n(x)$ of detected objects from a sample $x$ is \( \sum_{o}  \mathbbm{1} \{ \ f_{l_0}^o(x) > 0.5  \} \), where $\mathbbm{1}$ signifies the indicator function that takes the value one if the inequality holds and the value zero otherwise.

The goal of the object detector attack is to reduce the number of detected objects by modifying the minimum number of pixels, which is formally expressed as:

\begin{equation}
\begin{aligned}
\underset{\delta}{\min} \quad & {\max} \,   (n(x) - n(\overline{x})) \\
\text{s.t.} \quad & \left\| \delta \right\|_0 \leq \epsilon, \quad \epsilon \in \mathbb{Z}^+.
\end{aligned}
\end{equation}
where \(\epsilon\) denotes the attack level and \(\overline{x}\) denotes the perturbed image.
Therefore, our objective is to eliminate bounding boxes from the detection model by iteratively accumulating pixel attacks.
We address this problem by combining one-step REINFORCE\cite{REINFORCE} with our approach.

\subsection{Remember Process}
\label{Remember Process}
\paragraph{Agent.}
We construct an environment where an attacking agent interacts to generate adversarial images. 
The agent's policy utilizes a CNN-based architecture, where given a sample $x \in \mathbb{R}^{C\times H\times W}$, the agent observes the image and takes actions to determine the location (X, Y coordinates) and brightness (R, G, B) to modify, then generates perturbed images. 
We define two types of actions for brightness: "Write" and "Erase." 
The "Write" action overwrites the pixel with the maximum brightness, while the "Erase" action sets the brightness to zero.
This configuration is chosen because, based on our experience, the attack success rate is higher when applying maximum changes to the pixels. 
Figure \ref{fig:fig1} shows the adversarial images generated by the agent.
The agent generates the actions through random sampling of normal distributions, where the means and standard deviations are trained by the neural networks.
The set of actions $A_t$, where $t$ represents the training epoch of RL, contains $\mathcal{N}$ subsets corresponding to the number of attack pixels in each Remember process.
These subsets are composed of the X, Y coordinates, and brightness values for each channel. The set $A_t$ is defined as $ \{ \dashuline{a_{1}^1, a_{2}^1, a_{3}^1, \cdots, a_{C+2}^1}, \cdots, \dashuline{ a_{1}^\mathcal{N}, a_{2}^\mathcal{N}, a_{3}^\mathcal{N}, \cdots, a_{C+2}^\mathcal{N} } \} $, where $a_1$ and $a_2$ represent the X and Y coordinates, respectively, and $a_3$ to $a_{C+2}$ represent the brightness values for each channel. For the "Write" action, the brightness values are set to the maximum value, whereas for the "Erase" action, they are set to 0. 
For each pixel, the perturbed image $\overline{x}$ is generated as follows:
\begin{equation}
\overline{x}_{i,j,k} = 
\begin{cases} 
a_{i+2}^d & \text{if } j = a_{1}^d \text{ and } k = a_{2}^d, \\
x_{i,j,k} & \text{otherwise}
\end{cases}
\end{equation}
where $i$, $j$, and $k$ are indices for channel, height, and width, respectively.
In other words, $\overline{x}_{i,j,k}$ represents the brightness at position $(j,k)$. 
Adding a pixel to the image is repeated for $d$ from $1$ to $\mathcal{N}$.
Hence, the image is perturbed by $\mathcal{N}$ pixels. 
The equation describes the generation of $\overline{x}$ by repeatedly altering the brightness of each channel at the position $(a_{1},a_{2})$ in the given sample $x$. 
The agent trains by using the gradient of the reward and the log probability of the sampled actions.

\paragraph{Environment.} The environment evaluates the image generated by the agent and assigns a reward. The reward $r$ is defined as:
\begin{equation}
    r = \sum_{o=1}^{n} f^o_{l_0}(x)-f^o_{l_0}(\overline{x}) + \Omega(\overline{x}),
\end{equation}
where $n$ represents the number of detected objects in the image $x$ for object detection, while $\Omega(\overline{x}) = (n(x) - n(\overline{x}))$ signifies the number of removed objects after the adversarial attack. 
$l_0$ is the index for the confidence score of the detected object. 
Hence, the reward is defined as the sum of the differences in confidence scores for each object plus the number of objects removed. 

For classification, $l_0$ is the index for the correct class, and $n$ is set to 1. $\Omega(\overline{x})$ is set to 1 if the adversarial image generation is successful and 0 otherwise.
In essence, the reward is calculated as the sum of the differences in the probability of the correct class between the perturbed and original image, along with an additional component indicating whether the model successfully created an adversarial example.

\paragraph{Memory.} 
The role of memory is to save the best reward value and its corresponding perturbed images.
The stored information is also used to determine when the Forget process should start.
Without memory, RL models tend to identify universally shared vulnerabilities in the clean images provided to the victim model.
In contrast, our objective is to generate adversarial attacks regardless of these common vulnerabilities.
To minimize unnecessary queries that converge on such vulnerabilities, we have incorporated memory concepts into the RL approach.
In our approach, memory stores the maximum reward values \(r^*\) and their corresponding perturbed images \(\overline{x}^*\) by selectively saving the higher reward as
\(
    r^* = \max(r^*, r_{t})
\),
where \(r_t\) denotes the rewards given by the environments during the $t$-th training epoch.
% These stored values represent the highest reward found by the RL during its exploration, along with the corresponding perturbed image. 
After training each epoch of data, the algorithm checks whether the reward values have bounded.
We define the rewards as bounded if the following condition is satisfied:
\begin{equation}
    \frac{r_{t}-r^*}{r^*}<\eta
\end{equation}
where $\eta$ signifies the bound threshold. 
This equation indicates that the rate of increase in the reward stored in memory is less than $\eta$.
The convergence of rewards is defined as the rewards being bounded for a certain period, denoted as $T$.
Both $\eta$ and $T$ are hyperparameters.
If the reward converges, the Remember process ceases and the Forget process starts.

\subsection{Forget Process}
\label{forget process}
The goal of the Forget process is to reset the trained RL model and its memory, and to feed the image \(\overline{x}^*\) as a new input for the reset RL model. 
Additionally, the maximum \(L_0\) increase for the reset RL model, as it is determined by the number of reward convergences, attack pixels, and channels.
This process is implemented to prevent the agent from overfitting, which can hinder effective exploration of new inputs. 
The impact of memory and initialization is discussed in Section \ref{section:Ablation}.

\begin{table}
 \caption{\textbf{The results of adversarial attacks on the ImageNet dataset.} Each score represents the mean success rate of the attack, mean \(L_0\) norm and mean the number of queries. In terms of the success rate, a higher value signifies better performance, whereas for the \(L_0\) norm and the number of queries, lower values are indicative of superior performance. The best method is highlighted in bold.}
  \centering
  \begin{tabular}{c|c|c|c|c|c}
    \toprule
    
    Model    & Test accuracy   & Attack        & Succes rate $\uparrow$             & \(L_0\) $\downarrow$ & Query $\downarrow$   \\
    \midrule
    \multirow{4}{*}{VIT-B\cite{VIT}} & \multirow{4}{*}{81.07 \%} & OnePixel\cite{Onepixel}      & 9.3 \%\            & 15  & 1453   \\
    & & ScratchThat\cite{Scratch}                 & 40.9 \%\               & 420       & 9418     \\
    & & Pixle\cite{Pixel}                       & 51.4 \%\            & 286  & 728   \\
    & & \textbf{RFPAR(Ours)}          & \textbf{64.1 \%\ }  & 211& 613\\
    
    \midrule
    \multirow{4}{*}{ResNeXt50\cite{ResNeXt}} & \multirow{4}{*}{77.62 \%} & OnePixel\cite{Onepixel}      & 8.1 \%\            & 15  & 5100   \\
    & & ScratchThat\cite{Scratch}                 & 38.1 \%\            & 95   & 1400   \\
    & & Pixle\cite{Pixel}                       & 89.1 \%\            & 538  & 663   \\
    & & \textbf{RFPAR(Ours)}          & \textbf{95.3 \%\ }  & 138& 442\\

    \midrule
    \multirow{4}{*}{RegNetX-32GF\cite{RegNet}} & \multirow{4}{*}{80.62 \%} & OnePixel\cite{Onepixel}      & 12.3 \%\            & 15  & 1358   \\
    & & ScratchThat\cite{Scratch}                 & 60.6 \%\               & 427    & 8653   \\
    & & Pixle\cite{Pixel}                       & 73.7 \%\            & 276  & 705   \\
    & & \textbf{RFPAR(Ours)}          & \textbf{88.4 \%\ } & 164   & 484   \\

    \midrule
    \multirow{4}{*}{DenseNet161\cite{DenseNet}} & \multirow{4}{*}{77.14 \%} & OnePixel\cite{Onepixel}      & 14.1 \%\            & 15  & 1248   \\
    & & ScratchThat\cite{Scratch}        & 60.6 \%\            & 425   & 8367   \\
    & & Pixle\cite{Pixel}              & 82.3 \%\            & 243   & 625   \\
    & & \textbf{RFPAR(Ours)}          & \textbf{91.7 \%\ }            & 152   & 464   \\

    \midrule
    \multirow{4}{*}{MNASNet\cite{MNASNet}} & \multirow{4}{*}{73.46 \%} & OnePixel\cite{Onepixel}      & 14.2 \%\            & 15  & 1128   \\
    & & ScratchThat\cite{Scratch}        & 65.3 \%\            & 425   & 8828   \\
    & & Pixle\cite{Pixel}              & 83.7 \%\            & 240   & 607   \\
    & & \textbf{RFPAR(Ours)}          & \textbf{95.0 \%\  }          & 150   & 442   \\

    \midrule
    \multirow{4}{*}{MobileNet-V3\cite{MobileNet}} & \multirow{4}{*}{74.04 \%} & OnePixel\cite{Onepixel}      & 8.1 \%\            & 15  & 1461   \\
    & & ScratchThat\cite{Scratch}        & 51.8 \%\            & 420   & 9293   \\
    & & Pixle\cite{Pixel}              & 69.6 \%\            & 306   & 769   \\
    & & \textbf{RFPAR(Ours)}          & \textbf{86.6 \%\  }          & 213   & 596 \\

    \bottomrule
  \end{tabular}
  \label{tab:tab1}
\end{table}

\section{Experiments}
\label{experiment}
Section \ref{Experimental setting} details the dataset, evaluation metrics, victim models, and hyperparameters used in our experiments.
In Section \ref{classification attak result}, we evaluate our proposed attack on image classification by comparing it with previous attack methods.
Section \ref{object detection attack result} compares the performance of our method on object detection, varying the attack dimension (\(\alpha = 0.01\) to \(0.05\)), and compares the results with other query-based attacks.
In Section \ref{experiments on a larger scale}, we conduct experiments on the Argoverse dataset, which has larger image dimensions, and discuss the findings. 
Finally, Section \ref{section:Ablation} presents an ablation study on the memory and initialization components we introduced. 
Additional experimental results can be found in Appendix \ref{apendix1} and \ref{appendix2}.

\subsection{Experimental Details}
\label{Experimental setting}
\paragraph{Datasets, Metrics and Hardware.} 
\label{data and metric}
For image classification, we use the validation dataset from ImageNet-1K\cite{ImageNet}.
To reduce computational costs, we extract one correctly classified image per category from the victim model, resulting in a total dataset of 1000 images for adversarial attack attempts. 
We evaluate our methods with respect to different victim models by calculating the success rate, $L_{0}$ norm, and the number of queries. 
The \textit{success rate} represents the percentage of successful adversarial attacks out of the 1000 images, with higher values indicating better performance. 
The \textit{$L_0$} norm refers to the number of non-zero elements in perturbation $\delta$, with lower values indicating better performance. 
The number of \textit{queries} indicates how often the victim model is queried to generate an adversarial example, with fewer queries indicating better performance. 
The \textit{ATA} (ATtacked Area) refers to the proportion of pixels in the image that were attacked, a lower value indicates fewer changes.
For object detection, we use the 2017 validation set from the MS-COCO dataset\cite{MS-COCO} and Argoverse-1.1 validation set\cite{Argoverse}.
To facilitate comparison with PRFA \cite{PRFA} and GARSDC \cite{GARSDC}, we use mAP to evaluate the attacks. 
The mAP is calculated as the average over thresholds ranging from IOU = 0.5 to 0.95.
Additionally, \textit{RM} indicates the average percentage of objects removed from the clean image, while \textit{RD} refers to the decrease in mAP.
Both a lower mAP and a higher RM indicate greater success.
Lastly, we used an AMD Ryzen 9 5900X, RTX 3090TI, and 64.0GB of RAM, running on Windows 11 with CUDA version 12.1.

\paragraph{Victim Models.}
For image classification, we select six pre-trained models on the PyTorch platform as victim models: VIT\cite{VIT}, ResNeXt50\cite{ResNeXt}, RegNetX-32GF\cite{RegNet}, DenseNet161\cite{DenseNet}, MNASNet\cite{MNASNet}, and MobileNet-V3\cite{MobileNet}. We compare the performance of our attack with OnePixel\cite{Onepixel}, ScratchThat\cite{Scratch}, and Pixle\cite{Pixel}. For object detection, we use the pre-trained YOLOv8n model from the YOLOv8\cite{YOLOv8} platform and the pre-trained DDQ DETR-4scale model\cite{DDQ} from the MMDetection platform.

\paragraph{Hyperparameter.}
\label{hypeerparameter}

Our attack method utilizes four hyperparameters: the maximum number of iterations, the pixel attack rate $\alpha$, the bound threshold $\eta$, and the duration $T$ for maintaining the convergence condition. In the Remember process, $\alpha$ is a hyperparameter that determines the number of pixels to attack, proportional to the image size. The number of pixels $\mathcal{N}$ to be attacked is defined as $(H+W)/2\times\alpha$. By default, we set the maximum number of iterations to 100 and $\eta$ to 0.05. For image classification, we use $T = 3$ and $\alpha = 0.01$. For object detection, we experiment with $T = 20$ and $\alpha$ values ranging from $0.01$ to $0.05$.

\subsection{Evaluation of Classification Attacks}
\label{classification attak result}
Table \ref{tab:tab1} presents a performance comparison of various adversarial attack methods on different victim models for image classification. RFPAR consistently achieves the highest success rate, significantly outperforming the other three attack methods. 
For instance, for the VIT model, RFPAR achieves a success rate of 64.1\%, compared to OnePixel's 9.3\%, ScratchThat's 40.9\%, and Pixle's 51.4\%.
The trend is similar for other models, with RFPAR showing substantial improvements in success rate. Regarding the \(L_0\) norm, which measures the sparsity of the perturbations, RFPAR generally achieves a lower \(L_0\) norm than ScratchThat and Pixle but higher than OnePixel.
For example, in the case of ResNeXt50, RFPAR has an \(L_0\) norm of 138, compared to OnePixel's 15, ScratchThat's 95, and Pixle's 538. 
While OnePixel has the lowest \(L_0\) norm, its success rate is significantly lower than RFPAR's, indicating a trade-off between perturbation sparsity and attack effectiveness. In terms of the number of queries, RFPAR requires fewer queries than the other methods, except for OnePixel in some cases. 
This demonstrates that RFPAR is more efficient in terms of query cost, which is crucial for practical adversarial attacks. Overall, RFPAR exhibits superior performance across all victim models in terms of success rate while maintaining competitive \(L_0\) norms and requiring fewer queries compared to other methods, making it an effective and efficient approach.

\begin{table}
 \caption{\textbf{Attack Results on Object Detection Models.} 
 The subscripts after RFPAR denote a pixel attack rate, \(\alpha\). 
 RM indicates the average percentage of objects removed from the clean image. 
 $L_0$ represents the average $ \| \delta \|_0$. Query denotes the average number of queries made to the victim model. Higher RM, lower mAP, lower $L_0$, and lower Query values indicate better performance.
 }
  \centering
    \begin{tabular}{lccccccccc}
    \toprule
    \multirow{2}{*}{Attacks} & \multicolumn{4}{ c }{YOLOv8\cite{YOLOv8}} & & \multicolumn{4}{ c }{DDQ\cite{DDQ}}\\
    \cmidrule{2-5}
    \cmidrule{7-10}
    {}& RM $\uparrow$   & mAP $\downarrow$  & L$_0$ $\downarrow$  & Query $\downarrow$  & & RM $\uparrow$ & mAP $\downarrow$  & L$_0$ $\downarrow$  & Query $\downarrow$   \\
    \midrule
    clean        & - & 0.398 & - & -  & & - & 0.376 & - & -  \\
    RFPAR$_{0.01}$ &0.65 & 0.218 & 521 & 1403  & & 0.60 & 0.125 & 391      & 1450  \\
    RFPAR$_{0.02}$ &0.70 &0.187 & 955      & 1427  & & 0.73 & 0.103 & 787 & 1690  \\
    RFPAR$_{0.03}$ &0.75 &0.151 & 1459     & 1374  & & 0.76 & 0.075 & 1074 & 1512  \\
    RFPAR$_{0.04}$ &0.76 &0.150 & 1814     & 1348  & & 0.80 & 0.061 & 1429 & 1457  \\
    \textbf{RFPAR$_{0.05}$} &0.91 &0.111 & 2043     & 1254  & &0.83 &0.054 &1780      &1528  \\
    \bottomrule

  \end{tabular}
  \label{tab:tab2}
\end{table}

\subsection{Evaluation of Object Detection Attacks}
\label{object detection attack result}

\begin{wraptable}{r}{5.2cm}
\caption{\textbf{Comparison to other methods.} RD means reduction in mAP.}
\centering
\begin{tabular}{lcc}
\toprule
\multirow{2}{*}{Attacks} & \multicolumn{2}{ c }{YOLO}  \\
\cmidrule{2-3}
{}& RD $\uparrow$  & Query $\downarrow$    \\
\midrule
PRFA\cite{PRFA} & 0.21 & 2949 \\
GARSDC\cite{GARSDC} &\textbf{0.29} & 2691 \\
RFPAR &\textbf{0.29} &\textbf{1270}  \\
\bottomrule
\end{tabular}
\label{tab:tab3}
\end{wraptable}

Attacking object detection models is more challenging than attacking image classification models because there are more objects to consider in the object detection task. More pixels need to be modified, adjusted by \(\alpha\) from 0.01 to 0.05, to deceive the victim models.
Table \ref{tab:tab2} compares the performance of different \(\alpha\) values of the RFPAR method on two object detection models, YOLOv8 and DDQ.
The RM rate for YOLOv8 increases from 0.65 (RFPAR$_{0.01}$) to 0.91 (RFPAR$_{0.05}$) and for DDQ from 0.60 to 0.83, indicating that stronger attacks remove more detected objects. 
The mAP also decreases from 0.218 to 0.111 for YOLOv8 and from 0.125 to 0.054 for DDQ.
At $\alpha=0.05$, our attack successfully reduced the mAP by an average of 0.301 and achieved a RM of 0.87.
The number of queries remains relatively stable, ranging from 1254 to 1427 for YOLOv8 and from 1450 to 1690 for DDQ, suggesting a consistent query cost despite increasing perturbation intensity.
Overall, the results indicate that the RFPAR method is highly effective in generating adversarial attacks on object detection models, balancing perturbation sparsity, and attack effectiveness while maintaining query efficiency.

To demonstrate the effectiveness of our method, we compared it with other query-based black-box attacks. 
Table \ref{tab:tab3} shows the performance of three different attack methods - PRFA, GARSDC, and RFPAR - on the YOLO object detection model.
In this table, RD refers to the decreased mAP value, and Query indicates the average number of queries.
The RFPAR method shows strong performance by achieving the highest RD (tied with GARSDC) and requiring the fewest queries.
This indicates that RFPAR is not only effective in reducing the YOLO model's performance but also efficient in terms of the number of queries needed to achieve this reduction.
GARSDC also demonstrates high effectiveness with the same reduction as RFPAR but requires more than twice the number of queries.
Overall, RFPAR stands out as the most balanced and efficient attack method in this comparison.

\subsection{Experiments on a Larger Scale Data}
\label{experiments on a larger scale}

\begin{wraptable}{r}{7.5cm}
 \caption{\textbf{Comparison on dataset.} ATA means the ratio of altered pixels to the image size.}
  \centering
  \begin{tabular}{lcccc}
    \toprule
    \multirow{2}{*}{Datasets} & \multicolumn{4}{ c }{YOLO}  \\
    \cmidrule{2-5}
    {}& RM $\uparrow$ & RD $\uparrow$ & ATA $\downarrow$ &Query $\downarrow$ \\
    \midrule
    MS-COCO & 0.91 & 0.29 & 0.02 \% & 1270 \\
    Argoverse &0.94 & 0.05 & 0.10 \% & 1906 \\
    \bottomrule
  \end{tabular}
  \label{tab:tab4}
\end{wraptable}

To verify the effectiveness of our proposed method on larger dimensions 1920$\times$1200, we randomly selected one video sample from the Argoverse dataset and conducted experiments using YOLOv8. The experimental results are presented in Table \ref{tab:tab4}. The RM achieved 0.94, indicating a successful reduction in the number of detected objects.
Argoverse achieved a RM of 0.94, similar to the RM observed for MS-COCO.
The ATA for these datasets was 0.1\% and 0.02\%, respectively, indicating that only a very small portion of the image area was attacked.
However, the mAP did not decrease as significantly as in previous experiments. This discrepancy can be explained by considering that RFPAR primarily reduces the number of objects detected. If a particular class has many objects, reducing their number may not significantly impact the overall mAP due to the presence of other classes.
In summary, while RFPAR successfully removes objects in the larger Argoverse dataset, its effectiveness in reducing mAP is limited in datasets with a high density of objects in specific classes.

\subsection{Ablation study}
\label{section:Ablation}
\begin{figure}[h]
  \centering
  % \includesvg[width=\textwidth]{figure/ablation_modified.svg}
  \includegraphics[width=1\textwidth]{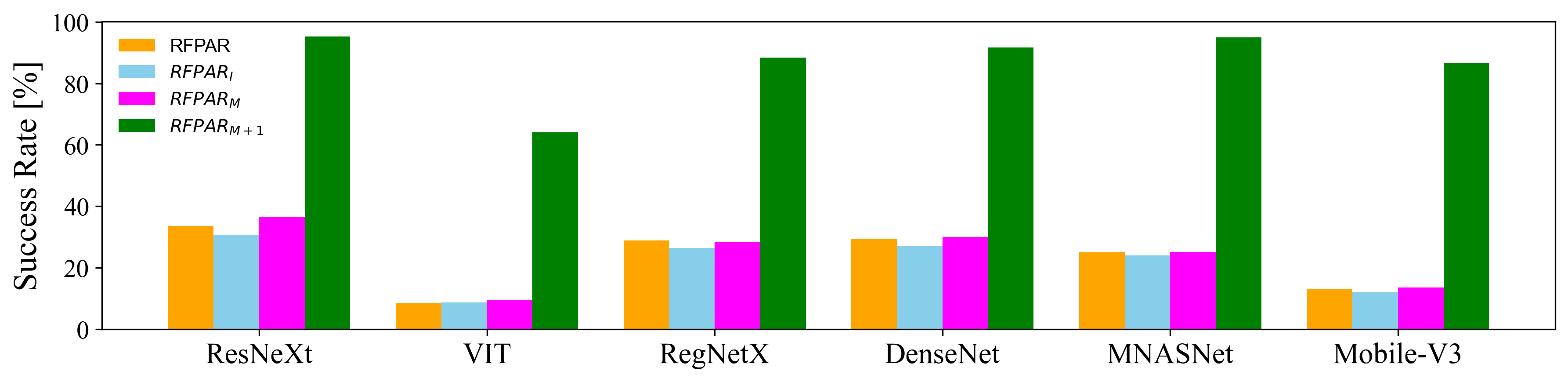} 
  \caption{\textbf{Ablation study.} The x and y axes show different victim models and the attack success rate, respectively. The notation $_I$ signifies the inclusion of the initialization step in the Forget process, and $_M$ denotes that the Remember process incorporates memory.}
  \label{fig:fig3}
\end{figure}

In this section, we analyze the impact of Initialization (I) and Memory (M) on our model's performance. 
If Initialization is ablated in the Forget process, the Agent is not reinitialized and retains information from the previous Remember process. On the other hand, if Memory is ablated, the Agent's reward, instead of the reward stored in Memory, serves as a bound condition. We conduct ablation experiments under similar query conditions and present the results in Appendix \ref{appendix5}.
As shown in Figure \ref{fig:fig3}, RFPAR denotes the baseline state without I and M, while RFPAR$_*$ indicates the inclusion of specific processes. Comparing RFPAR and RFPAR$_M$, it is evident that the introduction of memory significantly enhances the attack success rate. This result suggests that the RL method benefits from storing the highest-reward images of restricted pixels during each Forget process. When comparing RFPAR$_I$ and RFPAR$_{M+I}$, we observe that initialization prevents RL model from overfitting to specific patterns and escaping local optima, thereby improving performance. Conversely, the comparison between RFPAR and RFPAR$_I$ indicates that Initialization alone, without memory, has a negligible impact. This finding implies that RL model without memory fails to generate meaningful adversarial attacks. In summary, memory supports RL model in generating effective adversarial attacks, while Initialization prevents overfitting and enhances overall performance.

\section{Conclusion}
\label{conclusion}
In this paper, we propose the Remember and Forget Pixel Attack using Reinforcement Learning (RFPAR) for attacking neural network models with limited pixels. Traditional pixel-based attacks have been confined to image classification, but our method extends this approach to include object detection as well. For image classification, we compared the performance of RFPAR against OnePixel, ScratchThat, and PIXLE across six victim models using the ImageNet-1K dataset, and RFPAR demonstrated superior performance. In object detection, we evaluated RFPAR on the MS-COCO dataset using YOLOv8 and DDQ models, comparing it with PRFA and GARSDC attacks. RFPAR achieved performance comparable to the state-of-the-art query-based attack GARSDC, while reducing the number of queries by 52.8\%, proving its efficiency. Additionally, we showed that RFPAR is capable of performing pixel attacks on larger datasets, specifically the Argoverse dataset with dimensions, surpassing the sizes of the ImageNet and MS-COCO datasets.
Our findings may enable malicious individuals to compromise real-world AI systems. Consequently, research on defenses against adversarial attacks is becoming increasingly important.

\paragraph{Broader Impacts.} Defects in camera sensors, such as hot pixels or dead pixels, can impact image quality and degrade the performance of neural network models. Our approach mimics these camera defects. In this paper, RFPAR simulates real-world issues by replacing specific pixels with values of either zero or one, inducing incorrect predictions by the neural network. Since these types of perturbations can occur in practice, it is crucial for neural networks to be robust against them. However, research on pixel-based $L_0$ attacks is limited compared to other types of attacks. Our approach helps analyze model vulnerabilities with respect to both adversarial attacks and real-world scenarios, contributing to the development of more robust neural networks that can withstand such defects. Additionally, the phenomenon where the prediction changes with only a small number of pixel modifications that do not alter the overall meaning can be considered an anomaly in artificial neural networks. This type of attack provides important insights into understanding the limitations of neural networks.

\paragraph{Limitations.} In this work, the pixel values are either zero or one. While the meaning remains unchanged, this can still make it noticeable in certain cases. Additionally, the Forget process is quite simple. 
The time complexity of RFPAR is worse than that of other pixel attacks. However, at the ImageNet scale, RFPAR outperforms others in speed. This result is presented in Appendix \ref{appendix6}.
In future work, we will apply meta-learning to the Forget process and aim to reduce not only $L_0$, but also $L_{\infty}$.

\paragraph{Negative Impacts.} In applications like defective product detection \cite{defective} and disease prediction systems \cite{COVIDNetAT}, adversarial attacks could degrade product quality or lead to incorrect diagnoses, which may have serious, or even fatal, consequences. Our proposed approach increases the effectiveness of query-based black-box attacks, making them more applicable to real-world scenarios. As a result, vision AI systems may face significant threats to their functionality and reliability. Therefore, it is crucial for these systems to proactively identify potential vulnerabilities and implement robust defenses.

\paragraph{Mitigation of Risks.} Our method requires an average of over 1000 queries to successfully deceive an object detection model. Similarly, as shown in Table \ref{tab:varying iteration in transformer} in the Appendix \ref{appendix3}, transformer-based models also require an average of over 1000 queries to achieve a high success rate. If we limit the number of queries to around 1000 in a short period of time, our method can easily defend the model. For CNN-based models, since fewer queries are needed, limiting the queries to 400 can effectively defend the model. Additionally, according to the attack results on adversarially trained models shown in Table \ref{tab:adv} in the Appendix \ref{appendix4}, adversarial training effectively reduces the attack success rate and increases the number of queries needed. Therefore, by adversarially training the models and appropriately limiting the queries, this attack can be defended against.

\begin{ack}
This research was supported by the Korea Institute for Advancement of Technology (KIAT) through the Ministry of Trade, Industry, and Energy (MOTIE) of the Korean government (Grant No. P0017124, HRD Program for Industrial Innovation), the Korea Agency for Infrastructure Technology Advancement (KAIA) under the Ministry of Land, Infrastructure, and Transport (Grant No. RS-2022-00156364), and the Ministry of Education and the National Research Foundation of Korea as part of the "Convergence and Open Sharing System (NCCOSS)" Project.
\end{ack}

\medskip
\bibliographystyle{unsrt}
\bibliography{references_new}

% %%%%%%%%%%%%%%%%%%%%%%%%%%%%%%%%%%%%%%%%%%%%%%%%%%%%%%%%%%%%
\newpage
\appendix

\section{Related Work}
\paragraph{Adversarial Attack.}
Adversarial attacks manipulate clean images with imperceptible modifications to fool Deep Neural Networks (DNNs) into making incorrect predictions. These attacks are broadly divided into black-box and white-box attacks. White-box attacks leverage detailed information about the target model, including training data and gradients, to craft adversarial images. In contrast, black-box attacks, which do not rely on any internal information about the victim model, are divided into transfer-based and query-based strategies. Transfer-based attacks create adversarial images using a surrogate model, aiming for these examples to also be effective against the target model. Query-based attacks iteratively modify clean images and query the victim model, using the resulting confidence scores to refine the attack. Typically, attack strategies are evaluated using the $L_p$ norm to restrict the perturbation to remain imperceptible to humans\cite{fgsm, ffgsm, pgd, tpgd}.

\paragraph{Black-box Pixel Attack in Image Classification.}
Unlike other metrics, the $L_0$ norm, also known as pixel norm, targets only a small subset of pixels in a clean image rather than attacking all of them. The pioneering pixel attack method, OnePixel\cite{Onepixel}, employed Differential Evolution (DE) to generate adversarial images. An advanced approach, ScratchThat\cite{Scratch}, used DE to create curves and applied a parametric model to perturbations, reducing parameters and improving performance. A more recent study, PIXLE\cite{Pixel}, enhanced query efficiency and attack success rate by using a simple algorithm instead of DE. Briefly, PIXLE generates adversarial images by selecting arbitrary patches in a clean image and applying the brightness of these pixels to others. Although this method improved performance, it ignored pixel independence due to its reliance on patches and exhibited inefficiencies stemming from randomness in brightness mapping. The previous study, PatchAttack\cite{Patchattack}, utilized RL to embed textures in specific regions of the clean image, discovering vulnerable patches and reducing randomness, which significantly decreased the number of queries and improved attack success rates. Unfortunately, this method still depended on patches, requiring at least 3\% of the image area to be attacked. Our research focuses on eliminating patch dependency by attacking individual pixels and reducing randomness through RL. Extensive experiments demonstrate that our proposed attack outperforms the state-of-the-art methods in both query efficiency and attack success rate.

\paragraph{Query-based Adversarial Attack in Object Detection.} 
Adversarial attacks in object detection are more challenging than those in image classification. The first query-based attack in object detection, PRFA\cite{PRFA}, generates adversarial images using a parallel rectangle flipping strategy. Recent research, GARSDC\cite{GARSDC}, employs a genetic algorithm to create adversarial images, improving optimization efficiency by using adversarial examples generated from transfer-based attacks as the initial population. We extend our proposed attack from image classification to object detection. Experiments show that our method achieves a comparable mAP reduction on YOLO\cite{YOLOv8} to state-of-the-art methods while significantly reducing the number of queries, demonstrating its effectiveness in object detection.

\section{Theoretical Insight.}
We initially used a multi-step REINFORCE approach but identified issues, leading us to propose the Forget and Remember processes using one-step REINFORCE. Generating adversarial examples with multi-step REINFORCE involves the objective function $U = E \left[ \sum_{0}^{\mathcal{\tau}} \gamma^{\tau-t} R[s_t,a_t|\pi_\theta] \right]$, where $\gamma$ is the discount factor, $s_t$ is the image at step $t$, $a_t$ is the action at $s_t$, and the reward is $R[s_t,a_t|\pi_\theta] = f_{\theta,y}(s_{0}) - f_{\theta,y}(s_{t+1}),$ where $f_{\theta,y}$ is the confidence score of the true label y. Here, $a_t$ is a single pixel perturbation. We find that significant oscillations can be observed in the objective function. Let $\tau^*$ be the minimum number of steps to create an adversarial example. The sequence of pixels does not matter, leading to variations in the value of the objective function due to different orderings of $a_t$. Thus, for $i_t \in \{0,1,2,\cdots,\tau^*\} \text{ and } i_j \neq i_k$, the optimal objective function value is $U^* = E \left[ \sum_{0}^{\mathcal{\tau}} \gamma^{\tau-t} R[s_t,a_{i_t}|\pi_\theta] \right]$, with $\tau^*!$ permutations. This complicates training and increases the queries and $L_0$. To address this, we proposed the Forget and Remember process using one-step REINFORCE. Pixel perturbations at $\tau^*$ are defined as $A_{\tau^*} = \sum_0^{\tau^*} a_t$. By the intermediate value theorem, there exists a $C$ in $[x,x+A_{\tau^*}]$ such that $f_{\theta,y}(x) > f_{\theta,y}(C) > f_{\theta,y}(x+A_{\tau^*})$. We propose a Forget and Remember process using one-step REINFORCE to iteratively find this $C$, assuming $C \in \{x + a_0, x + a_1, \cdots, x + a_{\tau^*}\}$. This one-step approach avoids the fluctuations of multi-step methods, offering better query efficiency and lower $L_0$.
\newpage

\section{Experimental Results on Image Classification}
\label{apendix1}
\begin{figure}[h]
    \centering
    \begin{subfigure}[b]{0.45\textwidth}
        \centering
        \includegraphics[width=1\textwidth]{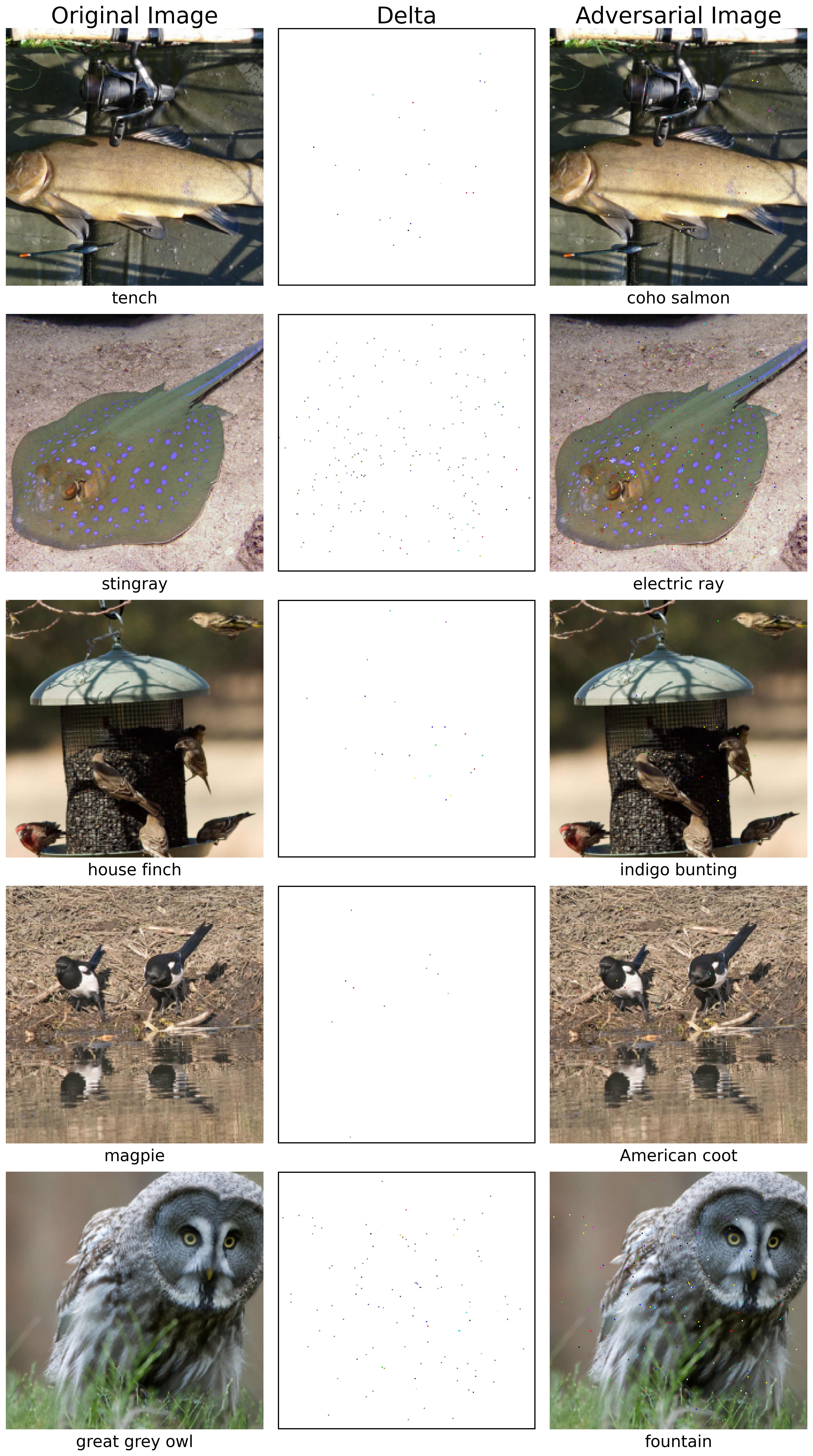} 
        \label{fig:subim1}
    \end{subfigure}
    \hfill
    \begin{subfigure}[b]{0.45\textwidth}
        \centering
        \includegraphics[width=1\textwidth]{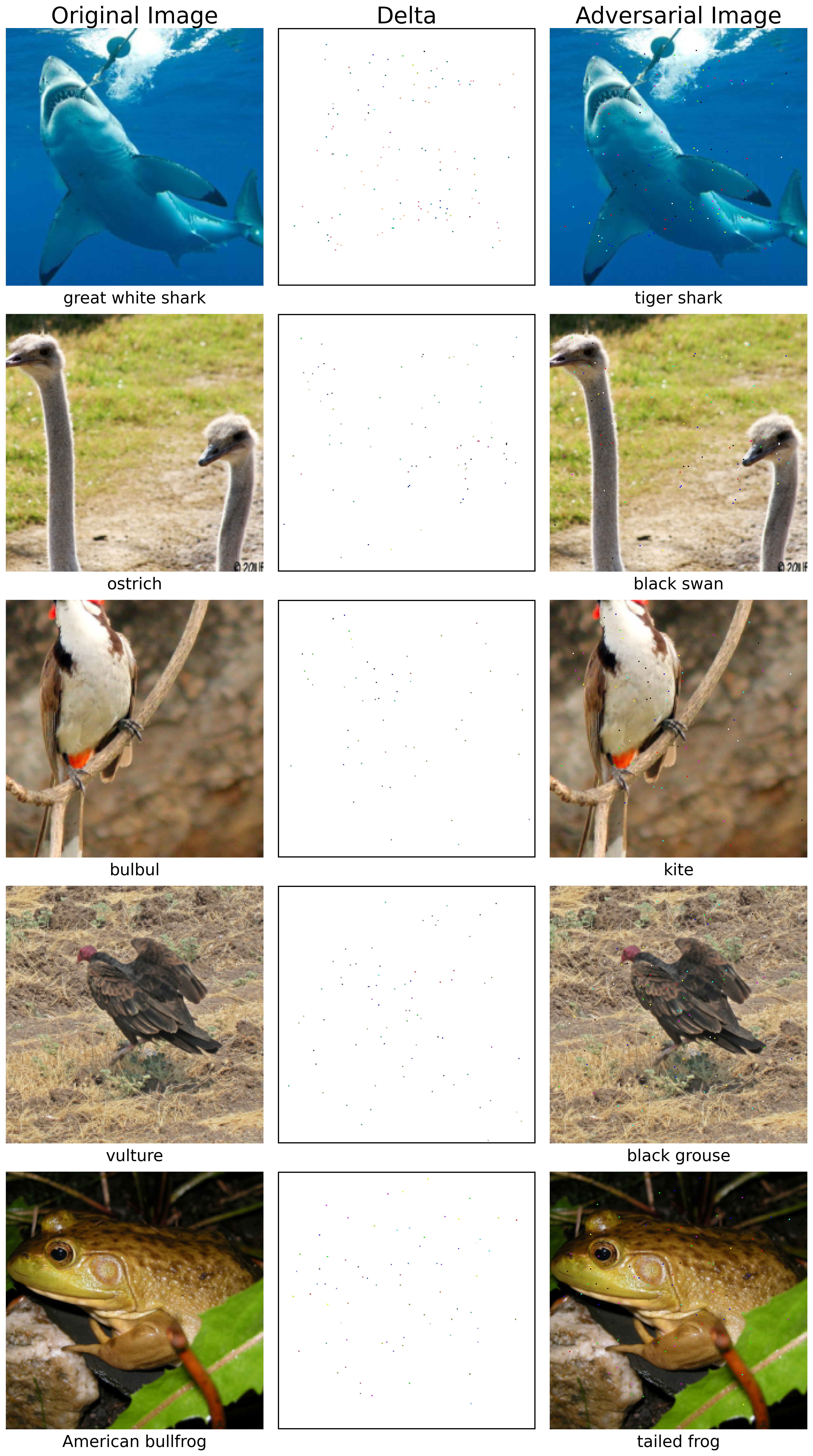}
        \label{fig:subim2}
    \end{subfigure}

    \caption{\textbf{Adversarial examples generated by RFPAR on the ImageNet dataset.} The "Original Image" is the original unaltered image, the "Delta" represents the difference between the Original Image and the Adversarial Image, and the "Adversarial Image" is the image with the altered prediction. The predicted labels are shown below the Original Image and the Adversarial Image.}
    \label{fig:app1}
\end{figure}

In this section, we present experimental results that could not be included in the main text. The results of attacking the ResNeXt50 model on the ImageNet-1K dataset are shown in Fig \ref{fig:app1}. The parameters for the attack were set as follows: the maximum number of iterations was 100, $\alpha$ was 0.01, $\eta$ was 0.05, and the duration $T$ was 3. 
\newpage
\section{Experimental Results on Object Detection}
\label{appendix2}
In this section, we present the experimental results for Object Detection from the main text. The results of attacking the YOLOv8n model on the MS-COCO dataset are shown, with the following parameters: the maximum number of iterations was set to 100, $\alpha$ ranged from 0.01 to 0.05, $\eta$ was 0.05, and the duration $T$ was 20. These results can be reproduced using the provided code.

\begin{figure}[h]  % 'h' stands for 'here'
   \centering
   \includegraphics[width=0.75\linewidth]{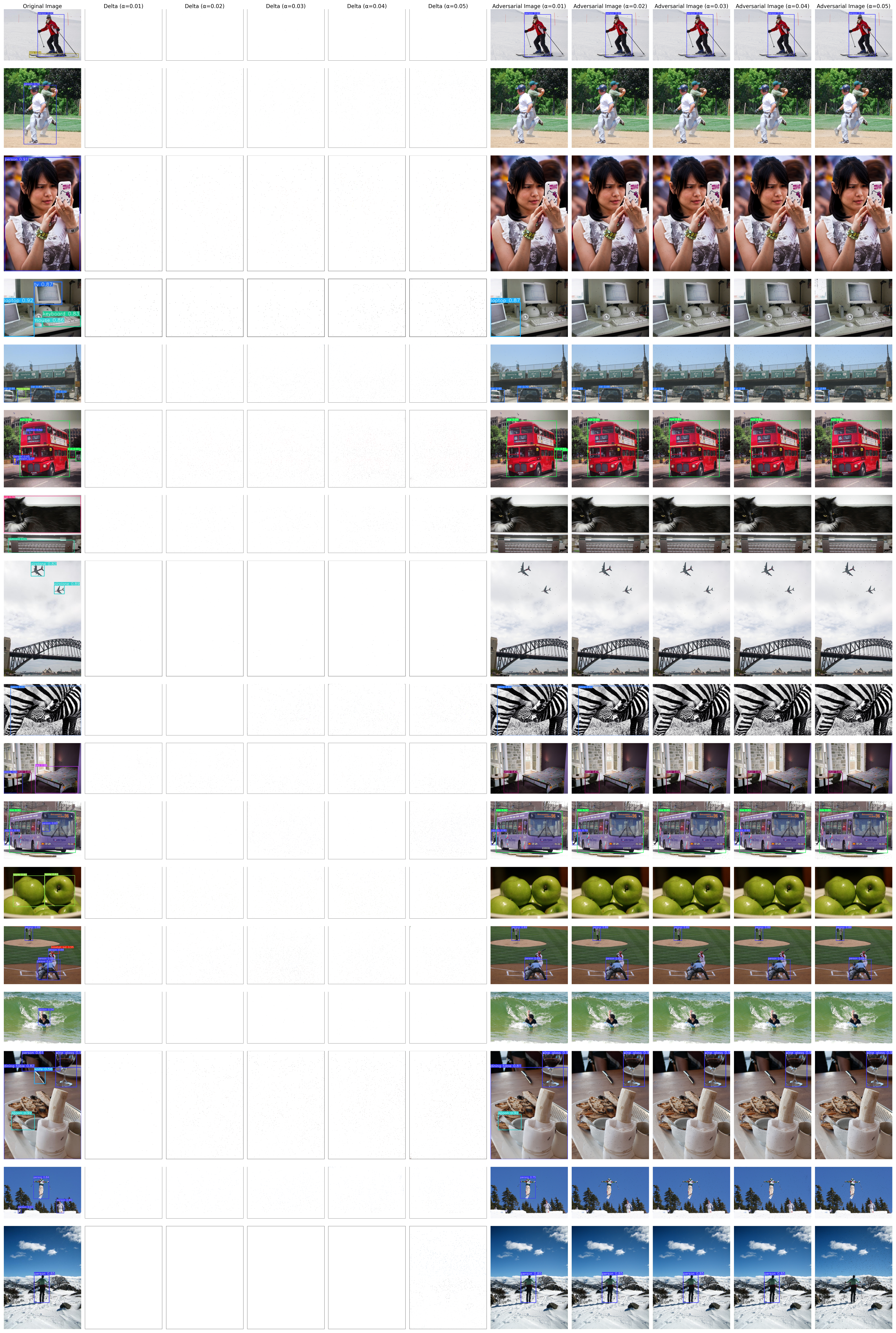}
    \caption{\textbf{Adversarial examples generated by RFPAR on the MS-COCO dataset.} The Original Image represents the unaltered image, and the Delta shows the difference between the Original Image and the Adversarial Image. The parameter $\alpha$ is a hyperparameter that determines the attack level; a higher value of $\alpha$  attacks more pixels. We conducted experiments with $\alpha$ ranging from 0.01 to 0.05. The Delta Image resulting from $\alpha$ values of 0.01 to 0.05 is presented in columns 2 to 6, and the Adversarial Image generated from the same $\alpha$ values is shown in columns 7 to 11. The Adversarial Image typically indicates an image with a changed prediction, but in this context, it also includes unsuccessful attacks. We present the results of Delta and Adversarial Images according to different values of $\alpha$.}
    \label{fig:app2}
\end{figure}

\newpage

\section{Additional Experiments}
\label{appendix3}
In this section, we present additional experiments that were omitted due to page limitations. Table \ref{tab:additional exp in transformer} shows the results on various transformer-based models (ViT-L \cite{VIT}, Swin-V2 \cite{swinv2}, and Deit-B \cite{deit}). These results demonstrate that RFPAR is effective on transformer-based models. Table \ref{tab:additional exp in OD} provides the results for object detection models, ATSS \cite{ATSS} and Deformable DETR \cite{DETR}, showing that RFPAR is also effective for object detection. Finally, Table \ref{tab:varying iteration in transformer} presents results for transformer-based models (ViT-B, L, H, Swin-V2, and Deit-B) with iteration limits of 100 and 200. These results indicate that RFPAR requires more queries to achieve a comparable attack success rate on transformer-based models compared to CNN-based models.

\begin{table}[H]
  \centering
  \caption{The results of transformer-based classifiers.}
  \begin{tabular}{l|ccc|ccc|ccc}
    \toprule

    \multirow{2}{*}{Attacks} & \multicolumn{3}{ c }{ViT-L } & \multicolumn{3}{ c }{Swin-V2 }& \multicolumn{3}{ c }{Deit }\\
    \cmidrule{2-10}
    {}& SR $\uparrow$   &  $ L_0 \downarrow$  & Query $\downarrow$  & SR $\uparrow$   &  $ L_0 \downarrow$  & Query $\downarrow$   & SR $\uparrow$   &  $ L_0 \downarrow$  & Query $\downarrow$     \\
    \midrule
    OnePixel        & 8.9\%& 15& 1654& 5.0\%& 15& 1686& 8.4\%& 15& 1137\\
    Pixle &66.4\%&531& 1396& 66.8\%& 1052& 1509& 71.0\%& 551& 1473\\
    RFPAR &\textbf{78.0\%}&355& 1042& \textbf{69.4\%}& 608& 1096& \textbf{84.3\%}& 412& 1161\\
    
    \bottomrule

  \end{tabular}
  \label{tab:additional exp in transformer}
\end{table}

\begin{table}[H]
  \centering
  \caption{The results of object detection models.}
  \begin{tabular}{lccccccccc}
    \toprule

    \multirow{2}{*}{Attacks} & \multicolumn{4}{ c }{Atss } & & \multicolumn{4}{ c }{Deformable DETR }\\
    \cmidrule{2-5}
    \cmidrule{7-10}
    {}& RM $\uparrow$   & mAP $\downarrow$  & L$_0$ $\downarrow$  & Query $\downarrow$  & & RM $\uparrow$ & mAP $\downarrow$  & L$_0$ $\downarrow$  & Query $\downarrow$   \\
    \midrule
    clean        & - & 0.227 & - & -  & & - & 0.339& - & -  \\
    RFPAR$_{0.01}$ & 0.74 & 0.048 & 491  & 1530        & & 0.61& 0.170& 333& 1466\\
    RFPAR$_{0.02}$ & 0.88 & 0.026 & 1025 & 1633       & & 0.69& 0.134& 512& 1502\\
    RFPAR$_{0.03}$ & 0.90 & 0.026 & 1357 & 1504    & & 0.72& 0.135& 869& 1488\\
    RFPAR$_{0.04}$ & 0.91 & 0.008 & 1666 & 1243     & & 0.76& 0.110& 1200& 1488\\
    RFPAR$_{0.05}$ & 0.92 & 0.006 & 2074 & 1288  & & 0.78& 0.073& 1274& 1335\\
    \bottomrule

  \end{tabular}
  \label{tab:additional exp in OD}
\end{table}

\begin{table}[H]
  \centering
  \caption{The performance of RFPAR on transformer-based models with different iteration limits}
  \begin{tabular}{lcccccccc}
    \toprule

    \multirow{2}{*}{Model} & \multicolumn{3}{ c }{maximum of Iteration = 100}& & \multicolumn{3}{ c }{maximum of Iteration = 200}\\
    \cmidrule{2-4}
    \cmidrule{6-8}
    {}& Succes rate $\uparrow$  & L$_0$ $\downarrow$  & Query $\downarrow$&& Succes rate $\uparrow$  & L$_0$ $\downarrow$  & Query $\downarrow$  \\
    \midrule
    ViT-B &64.1\% & 211& 613 & &\textbf{83.4\%} & 352& 995\\
    ViT-L &59.9\% & 209& 618 & &\textbf{78.0\%} & 355& 1042\\
    ViT-H &62.2\% & 166& 582 & &\textbf{73.5\%} & 229& 917\\
    Swin-V2 &46.2\% & 352& 611 &&\textbf{69.4\%}& 608& 1096\\
    Deit-B &60.2\% & 249& 676 &&\textbf{84.3\%}& 412& 1161\\
    \bottomrule

  \end{tabular}
  \label{tab:varying iteration in transformer}
\end{table}

\section{Experiments on Adversarially Trained Models}
\label{appendix4}
In this section, we present experiments on adversarially trained models (Adv. ViT \cite{ADV_vit} and Adv. ResNeXt101 \cite{ADV_res}). Table \ref{tab:adv} shows that RFPAR is effective on these models, although its success rate is lower compared to generally trained models. Proportional calculations indicate that RFPAR reduced ViT's performance from 69.10\% to 37.11\%, which, according to Appendix D of the Adv. ViT paper \cite{ADV_vit}, is more effective than CW20 (38.92\%), PGD-20 (37.96\%), and PGD-100 (37.52\%), but slightly less effective than AutoAttack (34.62\%). This demonstrates that our black-box attack, RFPAR, is nearly as effective as white-box attacks, despite having access to only limited information.

\begin{table}[H]
  \centering
  \caption{The results of adversarial trained models.}
  \begin{tabular}{lccccccc}
    \toprule

    \multirow{2}{*}{Attacks} & \multicolumn{3}{ c }{Adv. ViT } & & \multicolumn{3}{ c }{Adv. ResNeXt101 }\\
    \cmidrule{2-4}
    \cmidrule{5-8}
    {}& Succes rate $\uparrow$  & $L_0\downarrow$  & Query $\downarrow$ & & Succes rate $\uparrow$  & $L_0\downarrow$  & Query $\downarrow$  \\
    \midrule
    OnePixel & 2.9\%& 15& 2083& & 4.4\%& 15& 1102\\
    Pixle &34.0\%&780& 1912& & 42.5\%& 302& 769\\
    RFPAR &\textbf{46.3}\%&547& 1452& & \textbf{57.4}\%& 243& 626\\
    \bottomrule

  \end{tabular}
  \label{tab:adv}
  
\end{table}

\section{Query in Ablation study}
\label{appendix5}
In this section, we present Query regarding the ablation study and conduct experiments under similar conditions to ensure a fair comparison of each process.

\begin{table}[H]
  \centering
  \caption{Query for ablation study.}
  \begin{tabular}{lcccccc}
    \toprule
    & ViT-B&ResNeXt& RegNetX& DenseNet& MNASNet&MobileNet-V3\\
    \midrule
    RFPAR& 614& 529& 623& 534& 461& 548\\
    RFPAR$_{I}$&662&404& 444& 404& 364& 348\\
 RFPAR$_{M}$& 712& 889& 820& 723& 726&659\\
    RFPAR$_{M+I}$&613&442& 484& 464& 442& 596\\
    \bottomrule

  \end{tabular}
  \label{tab:query}
  
\end{table}

\section{Time complexity}
\label{appendix6}
Given the input dimension size $N$ and constants $K_i$: OnePixel has $O(K_1)$ complexity, ScratchThat has $O(N^2)$, Pixle has $O(K_2)$, RFPAR has $O(N)$, PRFAR has $O(K_3)$, and GARSDC has $O(N)$. For image classification tasks, RFPAR's linear increase in time complexity with image size is more favorable than ScratchThat's exponential increase but less so than OnePixel and Pixle. In object detection tasks, both RFPAR and GARSDC see linear increases in time complexity with larger images, making them less advantageous than PRFAR.

The higher time complexity compared to most other studies is a limitation of our research. However, RFPAR generates attacks using neural networks, similar to GARSDC, and benefits from the high performance of GPUs, allowing for faster computations despite the increased time complexity. We present the experimental times in Table \ref{time}.

To improve efficiency, we propose integrating our method with meta-learning. RFPAR involves the agent learning afresh on the image multiple times, which can mitigate overfitting but also results in unnecessary queries. Meta-learning could enable the agent to quickly adapt to new tasks, enhancing efficiency by learning more rapidly.

\begin{table}[h]
    \centering
    \caption{The experimental times in Table \ref{tab:tab1} of the main paper}
    \begin{tabular}{clllll}
    \toprule
        & ViT & RegNetX-32GF & MNASNet & DenseNet161 & MobileNet V3 \\
    \midrule
        OnePixel&       3h 2m   &   4h 36m  &   50m &   2h 52m  &   50m\\
        ScratchThat&    5d 12h 39m&   11d 11h 27m&   3d 19h 1m&   7d 6h 43m&   6d 11h 8m\\
        Pixle&          4h 48m&   8h 16m&   3h 3m&   13h 33m&   5h 14m\\
        \textbf{RFPAR}&          \textbf{1h 20m}&   \textbf{1h 20m}&   \textbf{20m}&   \textbf{47m}&    \textbf{29m}\\
    \bottomrule

    \end{tabular}
    \label{time}
\end{table}

\end{document}